%% file: acl2023.tex
\pdfoutput=1
\documentclass[11pt]{article}

\usepackage{ACL2023}

\usepackage{times}
\usepackage{latexsym}
\usepackage{pifont}
\usepackage{newunicodechar}
\usepackage{algorithm}
\usepackage{multirow}
\usepackage{array}
\usepackage{subcaption}
\usepackage{bbm}
\usepackage{enumitem}
\usepackage[noend]{algpseudocode}
\newunicodechar{✓}{\ding{51}}
\newunicodechar{✗}{\ding{55}}

\usepackage[T1]{fontenc}
\usepackage[utf8]{inputenc}

\usepackage{microtype}
\usepackage{amsmath}
\usepackage{graphicx}

\usepackage{inconsolata}

\title{Teaching-Assistant-in-the-Loop: Improving Knowledge Distillation from Imperfect Teacher Models in Low-Budget Scenarios} 

\author{Yuhang Zhou \\
    University of Maryland \\
    College Park, USA \\
  \texttt{tonyzhou@umd.edu} \\\And
  Wei Ai \\
  University of Maryland \\
    College Park, USA \\
  \texttt{aiwei@umd.edu} \\}

\begin{document}
\maketitle
\begin{abstract}
\input{1_abstract}
\end{abstract}

\input{2_introduction}
\input{3_related}
\input{4_method_yy}
\input{5_experiment}
\input{6_result}
\input{7_conclusion}
\bibliographystyle{acl_natbib}
\bibliography{anthology, custom}

\appendix
\input{8_appendix}

\end{document}

%% file: 1_abstract.tex
There is increasing interest in distilling task-specific knowledge from large language models (LLM) to smaller student models.
Nonetheless, LLM distillation presents a dual challenge: 1) there is a high cost associated with querying the teacher LLM, such as GPT-4, for gathering an ample number of demonstrations; 2) the teacher LLM might provide imperfect outputs with a negative impact on the student's learning process. To enhance sample efficiency within resource-constrained, imperfect teacher scenarios, we propose a three-component framework leveraging three signal types. The first signal is the student's self-consistency (consistency of student multiple outputs), which is a proxy of the student's confidence. Specifically, we introduce a ``teaching assistant'' (TA) model to assess the uncertainty of both the student's and the teacher's outputs via confidence scoring, which serves as another two signals for student training. Furthermore, we propose a two-stage training schema to first warm up the student with a small proportion of data to better utilize student's signal. Experiments have shown the superiority of our proposed framework for four complex reasoning tasks.
On average, our proposed two-stage framework brings a relative improvement of up to \textbf{$20.79\%$} compared to fine-tuning without any signals across datasets.
% }

%% file: 2_introduction.tex
\section{Introduction}
Large language models (LLMs) have demonstrated state-of-the-art (SOTA) performance across a wide spectrum of tasks, and their efficacy is primarily attributed to their substantial model size, enabling them to possess capabilities that smaller models lack \cite{brown2020language, raffel2020exploring, chowdhery2022palm, touvron2023llama}. 
%In-context learning (ICL), where feeding the LLMs with several exemplars and instruction, has become the most typical way to induce the knowledge inside the language models \cite{brown2020language}. 
%On complex reasoning or question answering (QA) tasks, researchers have developed several prompting methods, such as chain-of-thought (CoT) and ReAct for ICL, to elicit the language models' reasoning abilities by generating a trajectory with intermediate reasoning or action steps \cite{wei2022chain, yao2022react}. 
In most cases, a pre-trained LLM undergoes several specialized fine-tuning stages, such as instruction tuning, self-supervised tuning, and reinforcement learning with human feedback \cite{touvron2023llama, openai2023gpt4}, in order to excel at downstream tasks. 
Nevertheless, fine-tuning LLMs are challenging, primarily due to the demanding computational resources required \cite{wei2022chain, chowdhery2022palm} and the limited access to LLMs, such as GPT-4 and Claude~\cite{ouyang2022training, openai2023gpt4} prevents further fine-tuning. 

\begin{figure*}[!ht]
  \centering
  \includegraphics[width=0.9\textwidth]{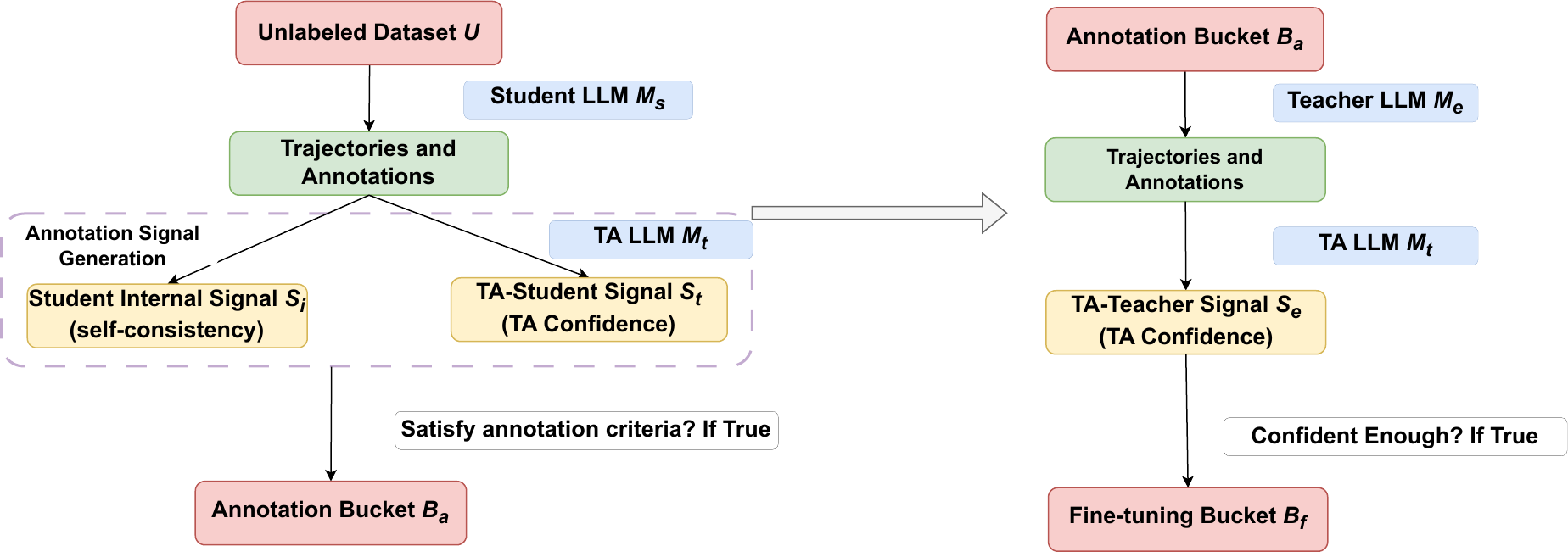}
   \caption{\label{fig:overview} \textbf{Overview of the proposed three-component KD framework.} The student model $M_s$ first makes the inference on the unlabeled dataset $U$ and we calculate student internal signal $s_i$ and TA-student signal $s_t$. Next, we distinguish whether this sample is worth annotating and add the satisfied ones to the annotation bucket $B_a$. Then, the teacher model $M_e$ will annotate the examples in $B_a$. If TA-teacher signal $s_e$ of the TA model is confident in the teacher's answer, we will add this question and annotations to the fine-tuning bucket $B_f$. 
   % [yy: can you change this png to a pdf? That will have a higher resolution. Also can we make this plot smaller?]
   }
\end{figure*}
%One of the major bottlenecks to scalably generalize the ICL for LLMs is the extremely large model size, causing high computational cost. Current state-of-the-art (SoTA) language models usually have hundreds of billions of parameters \cite{wei2022chain, chowdhery2022palm}, and some LLMs such as GPT-3.5 or GPT-4 are not open-source \cite{ouyang2022training, openai2023gpt4}. With these restrictions, deploying the LLMs to conduct the ICL at scale requires high computational requirements or costs a large expense for inference.

To resolve the above difficulties, knowledge distillation (KD) uses the outputs of a larger LLM (Teacher) to train a smaller model (Student), such as GPT-J-6B or LLaMa-7B \cite{gpt-j, touvron2023llama}. KD has gained significant attention and led to models such as Alpaca~\cite{alpaca2023} and Vicuna~\cite{vicuna2023}. 
The KD pipeline first utilizes In-Context Learning (ICL) on the teacher model to generate outputs, forming distillation sets and then use the teacher generations to fine-tune the student model.

However, current work of distilling LLMs presents two major difficulties: First, it can be prohibitively expensive to collect sufficiently large distillation sets, especially when querying proprietary LLMs like GPT-4, so a budget of asking the teacher is required in most real-life scenarios; Second, as LLMs may not have seen task-specific data, the quality of their demonstrations might be low. For instance, the zero-shot accuracy of InstructGPT on the Aqua task is only 34.25\% \cite{ouyang2022training}, and these suboptimal demonstrations negatively affect the student's performance, as elaborated in Section \ref{sec:experiments}. Effectively excluding flawed annotations, especially for unlabeled sets, is essential.

To address these KD challenges, we introduce a novel three-component KD framework for LLMs to learn efficiently from an imperfect teacher within budget constraints. 
In addition to the student and the teacher (as in standard KD pipeline), our framework introduces a Teaching Assistant (TA) model which mimics an TA in the real-life class and acts as an intermediary to communicate with students and teachers.
The TA can have a larger size and different architecture than the student, but smaller size than the teacher, leading to significantly reduced inference cost compared to the teacher. 

With the help of the TA model, our framework utilizes three types of new signals to refine the distillation. First, the student internal signal (self-consistency score) gauges the student's confidence and determines whether it should be ``forwarded'' to the teacher for annotation. Second, the TA model independently generates its TA-student signal for a question, aiding in the decision to annotate. Lastly, the TA model assesses annotations from the teacher (TA-teacher signal), deciding whether they merit inclusion in the student model's training dataset.

Moreover, the signal from the student model on a complex task is of lower quality for active selection. Therefore, we propose an extension of the two-stage training. Initially, we allocate 10\% of the annotation budget to fine-tune the student. After this ``warm-up'' stage, we utilize the remaining 90\% of the annotation budget. The student in both stages are fine-tuned within the proposed framework. Our experiments demonstrate that this approach further improves performance in various tasks.

\noindent Our work introduces three main contributions:
\begin{itemize}[leftmargin=*]
    \item 
    We introduce a TA model and develop a three-component KD framework that leverages three signals to improve the sample efficiency given a annotation budget and imperfect teacher LLM. 
    %In addition to the classical student-teacher knowledge distillation pipeline, we introduce a TA model to characterize the uncertainty of student answers and evaluate the correctness of the teacher annotations.
    %\item We develop a unified KD framework that utilizes the proposed TA models and other signals to actively select the samples for annotation and filter the noise in teacher annotations to improve the fine-tuning performance of the student model.
    \item 
    % \textcolor{red}{
    We further improve the three-component KD framework by introducing a two-stage training approach.
    % } 
    %splitting it into two stages and demonstrate the opportunity of curriculum training. 
    \item 
    We conduct extensive experiments on various datasets and public models and the proposed two-stage framework increases performance by up to 20.79\%, compared to fine-tuning without signals, suggesting the effectiveness of our framework. 
    %To quantify the effect of TA models and signals for student fine-tuning, we conduct extensive experiments on various datasets and multiple public models with the corresponding promoting methods.
    
\end{itemize}

%% file: 3_related.tex
\section{Related Work}
\subsection{Language model prompting}
Prompt-based learning refers to using prompts to induce the embedded knowledge in the language models to complete downstream tasks \cite{radford2019language, liu2023pre, raffel2020exploring}. Among various prompting methods, in-context learning (ICL) through adding a few demonstrations and instructions in the prompt to elicit the correct answer has been shown as an effective approach \cite{brown2020language}. Since the output from language models is sensitive to the instruction composition and demonstration selection, how to tune and design the prompts for ICL has attracted the attention of many researchers \cite{liu2023pre}. 
Chain-of-thought (CoT) prompting feeds the step-by-step examples and specifically designed instructions to ask the language models to generate the intermediate reasoning step of a complex task \cite{wei2022chain}. By grounding on the reasoning steps, the language models perform better on a wide variety of complex tasks. Based on CoT methods, researchers also divide the reasoning process into multiple steps \cite{creswell2022faithful, creswell2022selection}. Moreover, \citet{yao2022react} propose another prompting method to integrate the additional action step to obtain external knowledge, which has been shown to be more effective in reasoning and decision making tasks.

\subsection{Signals and feedback in language models}
Besides asking the LLM to generate more intermediate steps, some works utilize the signals or feedback generated from the language model output to continually refine the prompts and resolve the hallucination \cite{madaan2023self, shinn2023reflexion, diao2023active, su2022selective, peng2023check, welleck2022generating, liu2023aligning, li2024cfmatch}. \citet{diao2023active} and \citet{su2022selective} use the measured uncertainty or score to actively select the uncertain exemplars on the prompts to increase ICL performance. 
%Some other works rely on the language model itself to compose the natural language feedback to the output. They add the generated feedback and the last output to the prompt to avoid making the same mistake \cite{madaan2023self, shinn2023reflexion}. 
\citet{madaan2023self, shinn2023reflexion} rely on the language model itself to compose the natural language feedback. 
These signal-based methods are usually utilized in language model prompting (ICL) without any model parameter updates, which are distinct from our work since we designed signals for collecting high-quality examples for model finetuning. 
% In addition to ICL without any model updates, there is another line of work for language model is to distill the knowledge from the LLMs to small models via fine-tuning small models on the larger model generations.

\subsection{Knowledge distillation in language models}
% \textcolor{red}{
Knowledge Distillation (KD) can be framed as training small student models based on data generated from large teacher models to reduce model size and expense while retaining capabilities \cite{sanh2019distilbert, hinton2015distilling, buciluǎ2006model}.
For the KD pipeline with black-box LLMs, there are two lines of work. The first is to ask teacher models to generate the final answers and to do standard fine-tuning on the final answers \cite{yoo2021gpt3mix, schick2020exploiting, schick-schutze-2021-just, zhou2023scalable}. Another line of work is to ask the student model to fine-tune the teacher-generated CoT or other prompting trajectories (rationales) \cite{yao2022react, ho2022large, he2023teacherlm, shridhar2023distilling, hsieh2023distilling, wang2023making, feng2023citing}.
Through the developed prompting methods, researchers find that sequence-level distillation of reasoning steps generated by the teacher is more effective \cite{yao2022react, ho2022large}, enabling the reasoning or decision-making abilities in students. 
%Our work is within the scope of the sequence-level KD work, fine-tuning student models on the teachers' generations. We restrict the usage of the teacher model to limited and low-budget scenarios and assume that we cannot access the ground-truth labels of the training set. To solve these challenges, we utilize various signals from both student and teacher sides to obtain better sampling efficiency and data quality. 
However, most sequence-level KD works utilize the known correct or incorrect trajectories to fine-tune the student \cite{li2023turning, wang2023making, chen2023mcc, liu2023mind, an2023learning}. Moreover, they consider the scenario with sufficiently large fine-tuning set while the fixed teacher budget and the flawed teacher annotations in the unlabeled data are yet to be explored. 
\citet{yu2023metamath, liang2023mint} boost the fine-tuning set by rephrasing noisy rationales but do not actively select annotated samples to save teacher budget.
\citet{mirzadeh2020improved, son2021densely} also leverage Teacher Assistant (TA) models for KD tasks but within the realm of computer vision, diverging from our approach that applies TA models in sequence-level KD tasks.
Table \ref{table:related_work} summarizes the similarity and uniqueness of our work with previous related ICL or KD approaches.
% }

\begin{table}[t]
\small
\centering
\resizebox{\columnwidth}{!}{%
\begin{tabular}{lccccl}
\hline
Method           & \begin{tabular}[c]{@{}c@{}}Model \\ Updates\end{tabular} & \begin{tabular}[c]{@{}c@{}}Selective \\ Samples\end{tabular} & \begin{tabular}[c]{@{}c@{}}Teacher \\ Usage\end{tabular} & \begin{tabular}[c]{@{}c@{}}w/o \\ GT Labels\end{tabular}\\ 
\hline 
% & Reference              \\ \hline
Zero-shot~\cite{radford2019language}        & ✗                                                        & ✗                                                            & ✗                                                        & ✓          \\                                                       %& \cite{radford2019language} \\
Few-shot CoT~\cite{wei2022chain}      & ✗                                                        & ✗                                                            & ✗                                                        & ✗                 \\                                               % & \cite{wei2022chain}     \\
Active prompting~\cite{diao2023active} & ✗                                                        & ✓                                                            & ✓                                                        & ✓                   \\                                        %     & \cite{diao2023active}    \\
Finetune CoT~\cite{ho2022large}     & ✓                                                        & ✗                                                            & ✓                                                        & ✗  \\                                        
Distilling step-by-step~\cite{hsieh2023distilling} & ✓                                                        & ✗                                                            & ✓                                                        & ✗  \\
MetaMath~\cite{yu2023metamath} & ✓                                                        & ✗                                                            & ✓                                                        & ✗  \\
Finetune ReAct~\cite{yao2022react}  & ✓                                                        & ✗                                                            & ✓                                                        & ✗                          \\
%                                   & \cite{yao2022react}     \\
TA-in-the-loop (our work)         & ✓                                                        & ✓                                                            & ✓                                                        & ✓\\
\hline 
%                &                        \\ \hline
\end{tabular}
}
\caption{\label{table:related_work}  A comparison of our work to closely related prior ICL or KD approaches for LLMs.}
\end{table}

%% file: 4_method_yy.tex
\section{Methodology}
Figure \ref{fig:overview} shows the overview of our framework. Our three-component framework consists of two pipelines: 1): Collect a fixed budget of examples that meet annotation criteria and add them to the bucket $B_a$: the intuition here is that we want to choose the challenging examples which the student model has a chance to learn correctly instead of choosing examples randomly that will include oversimple or complicated examples beyond the student's learning ability 2): Annotate the examples from $B_a$ and return a high-quality fine-tuning set $B_f$: we want to filter out low-quality or wrong annotations generated by the teacher LLM. 

We leverage the interactions among the student model $M_s$, the teacher model $M_e$, and the TA model $M_t$ to generate three types of signals and achieve the above goals. For a given example, we will use the student internal signal $s_i$ as well as the TA-student signal from the TA model $s_t$ to decide whether the example should be added to $B_a$. After collecting the fixed budget of samples for $B_a$, we ask the teacher model to annotate those examples and apply the TA-teacher signal $s_e$ from the TA model to filter out low-quality demonstrations. 
%Our pre-trained student model $M_s$ will first do the few-shot ICL inference on a large unlabeled dataset $U$ and then calculate the student internal signal $s_i$ from the student-generated answer set or the external signal $s_t$ from the judgement of the TA model $M_t$. With the student signals, we distinguish whether this sample is worth annotating by the pre-defined criteria for each type of student signal. If the signal of a sample meets the criteria, we will add it to the annotation bucket $B_a$. After active selection, we ask the teacher model $M_e$ to again use the few-shot ICL to annotate the examples in the annotation bucket. Then the TA model $M_t$ labels the confidence level of the teacher-generated answers. If TA model is confident with the teacher's answer of a question, we will add this question and teacher annotations to the fine-tuning bucket $B_f$. Finally, we fine-tune our student model on the fine-tuning bucket and finish the knowledge distillation. 
We introduce these three signals in detail below, and the algorithms outlining this procedure are also presented in Algorithm ~\ref{alg:ta}.

\begin{algorithm}[t!]
\caption{\textbf{Three-component Knowledge Distillation Framework}}
\label{alg:ta}
\begin{algorithmic}[1]

\State \textbf{Input:} Unlabeled dataset $U$, student model $M_{s}$, TA model $M_{t}$, teacher model $M_{e}$, confidence prompt $P_{c}$, annotation criteria annotate$(\cdot)$, confidence set $C$, annotation bucket $B_{a}$, fine-tuning bucket $B_{f}$
\State \textbf{Output:} The desired fine-tuning bucket $B_{f}$

% \vspace{0.15cm}
% \Statex \textbf{\sf \textcolor{gray}{/* STEP 1. Check if input is worth annotating by collecting the student signals */}}
\For{each input $x$ from dataset $U$}
        \State $t_{i}(x) =M_{s}(x, P_i) \text{ for $i\in\{1,\cdots, n\}$ }$
        \State $s_{i} (x)$ = $\text{Uniq}(t_{1}(x) \dots t_{n}(x))$
        \State $s_{t} (x)$ = $M_{t}({x, t_1, P_c})$
        \If{$\text{annotate}(s_{i}, s_{t})$ is True} 
            \State Add $x$ into annotate bucket $B_{a}$
        \EndIf
\EndFor

% \Statex \textbf{\sf \textcolor{gray}{/* STEP 2. Ask teacher to annotate and use TA to filter the unconfident ones*/}}
\For{each input $x$ from dataset $B_{a}$}
        \State $t_{e}(x)$ = $M_{e}(x, P_{i})$
        \State $s_{e}(x)$ = $M_{t}(x, t_e, P_{c})$
        \If{$s_{e}$ in $C$} 
            \State Add $x$ into fine-tuning bucket $B_{f}$
        \EndIf
\EndFor

\State \Return{$B_{f}$}
\end{algorithmic}
\end{algorithm}

\subsection{Signals for annotation}

We propose two types of signals to decide whether an example meets the annotation criteria: student internal signal $s_i$ and TA-student signal $s_t$. 

%We will continuously add the questions satisfying the specified criteria to the annotation bucket $B_a$ until the data size in $B_a$ reaches the predefined budget.

\subsubsection{Student internal signal $s_i$}
Motivated by self-consistency work \cite{wang2022self}, we utilize the disagreement number (self-consistency number) as the measurement of the uncertainty of student models. For a given sample $x$ and prompt $P_i$, we use stochastic temperature sampling with a fixed temperature and repeat the process for $n$ times with answers $t_1, \cdots t_n$:
\begin{equation*}
    t_{i}(x) =M_{s}(x, P_i) \text{ for $i\in\{1,\cdots, n\}$ }
\end{equation*}

\noindent then the internal signal can be calculated by counting the unique values among $n$ answers
\begin{equation*}
    s_{i}(x)=\text{Uniq}(t_1(x),\cdots,t_n(x))
\end{equation*}
where Uniq is counting the unique answers from $t_1$ to $t_n$, and $s_i\in\{1, 2,\cdots, n\}$. In Table \ref{table:disagreement_examples} in the Appendix, we present two examples with different numbers of student disagreement.

%For our following experiments, we choose the hyperparameter $n=5$. 

\subsubsection{TA-Student signal $s_t$}
\label{sec:student_external}
We also utilize the TA model as an auxiliary signal to characterize the uncertainty of student's generations, given the observation that the confidence estimated by a LLM itself is prone to be overconfident \cite{diao2023active, si2022prompting}. For a given example $x$ and the student annotation $t$ (rationales and answer), the TA model is provided with a crafted confidence prompt $P_c$ to classify the student annotation $t$ 
\begin{equation*}
    s_t(x) = M_t(x, t, P_c)
\end{equation*}
where $s_t$ is a categorical variable belonging to the confidence set \{very confident, confident, not confident, wrong answer\}. Details and examples of the confidence prompt $P_c$ and $t$ are shown in Table \ref{table:ta_conf_prompt} and \ref{table:examples_ta_conf} in Appendix.

%Besides the internal signal from the student model, we can also utilize LLM to estimate the uncertainty of the student model, which can be obtained by querying the student model itself for confidence in its own generated answers. However, from observations in previous work, the confidence estimated by LLM itself is prone to be overconfident \cite{diao2023active, si2022prompting}. To solve this drawback, we introduce another LLM, namely the teaching assistant (TA) model $M_t$, which is larger and has more capabilities in complex tasks. After obtaining the answer $t_{s_0}$ from the student model, we provide the TA model with the answer, as well as a crafted confidence prompt $P_c$ containing four confident choices: (a) very confident, (b) confident, (c) not confident, and (d) wrong answer. Then we combine the ``confidence prompt" $P_c$ and the student's output $t_{s_0}$ and feed the concatenation to the TA model $M_t$ to generate the confidence choice $s_{t} = M_{t}(P_{c} \vert\vert t_{s_{0}})$. \textbf{TA-confidence} choice $s_t$ is a categorical variable from the confidence set \{very confident, confident, not confident, wrong answer\}. Details and examples of the confidence prompt $P_c$ and $t_{s_0}$ can be found in Table \ref{table:ta_conf_prompt} and \ref{table:examples_ta_conf} in Appendix.

\subsubsection{Annotation criteria}
\label{sec:annotation}
Our annotation criteria are formulated by the student internal signal $s_i$ and TA-student signal $s_t$. We compute the complexity score of an example $x$:
\begin{equation}\label{eq1}
    c(x) = \alpha\mathbbm{1}_{s_i(x)\in C_1} + \beta\mathbbm{1}_{s_t(x)\in C_2}
\end{equation}
where $\alpha,\beta\in\{0,1\}$ are the signal weights. 
% \textcolor{red}{
When $\alpha=1$ and $\beta = 1$, it means both signal $s_i$ and $s_t$ are utilized, and conversely if either $\alpha$ or $\beta$ is set to 0, the corresponding signal is not employed.
% } 
$\mathbbm{1}$ is the indicator function and $C_1, C_2$ are complexity sets to actively select the examples that bring the deepest learning curve for the student model. The example will be added to the annotation bucket $B_a$ if $c(x) \geq \alpha + \beta$ until the predefined budget is met. 

% \textcolor{red}{
In our experiment setup, we use $n=5$, $C_1=\{2,3\}$ and $C_2=$ \{confident, not confident\} after the preliminary exploration about hyperparameter combinations (Appendix \ref{sec:pre_annotation}), which ensures that we can choose questions that are not too easy or too hard to learn for the student model. 
% }

%When using the student internal signal to actively select samples for annotation, we put the samples with disagreement number $s_i$ equal to 2 or 3 ($\max(s_i) = 5$) in the annotation bucket $B_a$. The intuition of not selecting the most uncertain questions ($s_i$ = 4 or 5) is that even after fine-tuning, the student model cannot correctly complete the hardest questions for challenging tasks. For example, from the previous work, the final fine-tuning accuracy of a 6.7B student model on most reasoning tasks cannot exceed 70\% accuracy \cite{ho2022large}. Therefore, our strategy is to choose the uncertain samples, but not the hardest, in the annotation bucket $B_a$. 

%Similarly, when using the student external signal, we choose questions with TA-confidence $s_t$ equal to ``confident'' or ``not confident'' in the annotation bucket $B_a$. We will stop actively selecting after the number of questions in the annotation bucket reaches the predefined budget.

%We also consider using both two signals at the same time to select the questions for annotation, but the number of questions satisfying both criterion of two signals is small, e.g. only about 1/8 of the whole dataset for the HotpotQA task \cite{yang2018hotpotqa}, causing low sampling efficiency. Therefore, in extensive experiments, we will show two framework variants with either of the student signals and show the effectiveness of the student signals.

\subsection{TA-Teacher signal $s_e$}
%\subsection{Signal $s_e$ for filtering out low-quality annotations 
% [yy: this needs to be shortened]}
With the desired annotation bucket, we utilize the teacher model $M_e$ to annotate the question $x$ in the annotation bucket $B_a$ with the few-shot ICL. To verify the correctness of teacher annotation $t_e$, we apply the similar TA-confidence as discussed in Section \ref{sec:student_external}. The process of adding TA-confidence can be formulated as
\begin{equation*}
    t_e(x) = M_{e}(x, P_i), \quad s_{e}(x) = M_{t}(x, t_e, P_{c})
\end{equation*}
where $s_e$ is the TA-confidence of teacher annotations and serve as the TA-teacher signal.
We calculate the confidence score for an example by 
\begin{equation}\label{eq2}
    d(x) = \mathbbm{1}_{s_e(x)\in C_3}
\end{equation}
where $C_3$ are the confidence set. Teacher annotations will be added to the fine-tuning set $B_f$ only if $d(x)\geq 1$. 
In our experiments, we set $C_3=\{\text{very confident, confident}\}$. Note that our signal does not require an additional cost to call the teacher model, which makes our approach different from \cite{madaan2023self, shinn2023reflexion} where they propose filtering out low-quality samples by self-refine from the teacher model. 
%Instead of asking the teacher model to self-refine the annotations \cite{madaan2023self, shinn2023reflexion}, we choose to apply the low-cost TA model to examine the annotations; otherwise, the teacher's self-evaluation will cost a large proportion of limited budgets. 
%For a given sample $x$, if $s_e(x) \in \{\text{very confident, confident}\}$, we put it in the fine-tuning bucket $B_f$ to maintain high data quality.
% Unlike the selection criteria for student inference, we choose questions and annotations with TA-confidence equal to ``very confident'' or ``confident'' into the fine-tuning buckets $B_f$ and high confidence from the TA model will reduce noise in the fine-tuning bucket. 
Examples of questions, annotations, and the TA signal $s_e$ are shown in Table \ref{table:confidence_examples} in Appendix.

\subsection{Fine-tune the Student Model}
After collecting the fine-tuning dataset $B_f$, we fine-tune the student model with questions and teacher-generated annotations from the fine-tuning bucket. We apply the autoregressive language modeling as well as the cross entropy loss, which is the same as the pretraining objective to fine-tune the student model \cite{gpt-j}.

\begin{figure}[t]
  \centering
  \includegraphics[width=\linewidth]{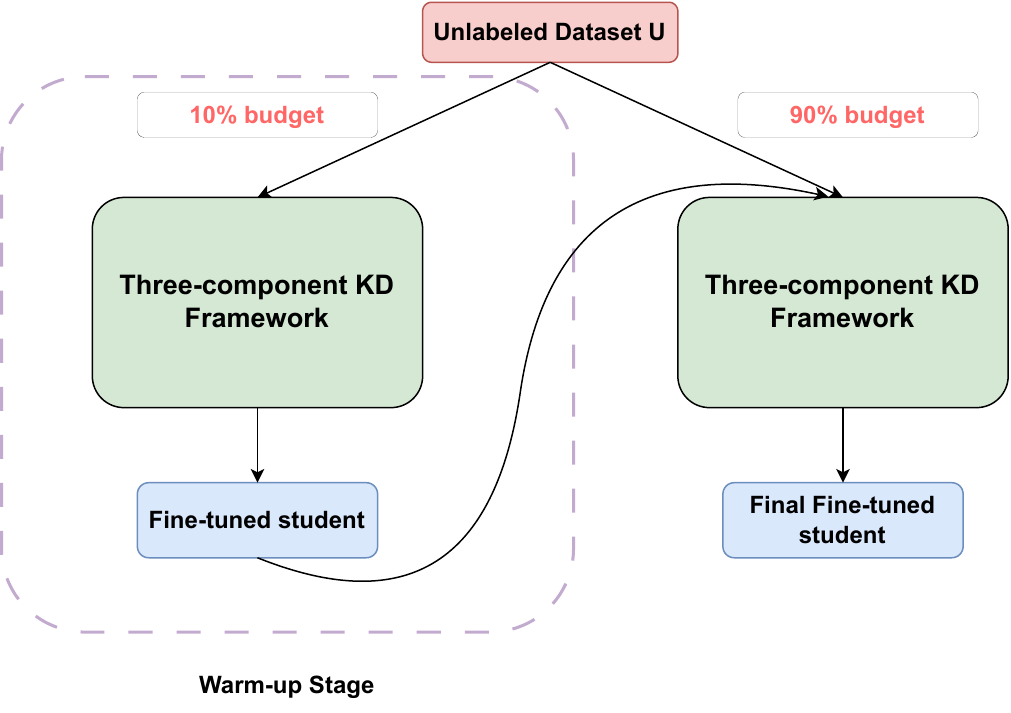}
   \caption{\label{fig:two_stage} \textbf{Overview of two-stage training extension.} We first use 10\% budget to fine-tune the student model and then use the other 90\% to further training the student. Both stages use signals to enhance the fine-tuning.}
\end{figure}

\subsection{Extension: Two-stage Training}
\label{sec:extension}
%To better utilize the internal signals from the student model, we propose the two-stage training extension to first utilize a small proportion of the training budget to warm up the student and then continue fine-tuning with the remaining budget in the second stage. The motivation behind this extension is that the student model's few-shot ICL can only obtain low accuracy for some tasks, so the uncertainty measurement from the student signal is not reliable. For example, as presented in Section \ref{sec:results}, for HotpotQA task \cite{yang2018hotpotqa}, the accuracy of the student's inference (GPT-J-6B) is only 5.2\%. With such a low accuracy, it is questionable whether student signals can effectively characterize the uncertainty of the student model. Moreover, even if the student signals are reliable, the samples extracted by signals are quite uncertain before fine-tuning, but may not be challenging enough for the student models fine-tuned halfway. 
While the aim of using student internal signal $s_i$ is to actively select the examples not too simple or challenging for the student to learn, the student answers $t$ for calculating $s_i$ could be quite inaccurate, for example, the accuracy of using GPT-J-6B as the student is only 5.2\% at HotpotQA task \cite{yang2018hotpotqa}. With such a low accuracy, two challenges will emerge. First, the signal $s_i$ cannot precisely reflect the difficulty and confidence of the student in the example. Second, the ratio of examples with $s_i(x) \in C_1$ is small (most examples with $s_i=4 / 5$), causing a low sampling efficiency.

To address these challenges, as shown in Figure \ref{fig:two_stage}, we develop a two-stage training. The first stage is the warm-up training, and we use 10\% of the budget to fine-tune the student with our framework (using signals).
% [yy: do we have TA model here? The graph has TA in the loop] 
For the second stage with a more proficient student model, we use 90\% of the budget to continue fine-tuning the student model with our framework. We expect that the student signals collected in the second stage are more reliable than the first stage, leading to a more performance boost.

%% file: 5_experiment.tex
\section{Experiments}
\label{sec:experiments}
Through our extensive empirical analysis, we aim to address the following research questions:
\begin{itemize}[leftmargin=*]
    \item \textbf{RQ1}: How efficient is our three-component framework compared to a classical KD method?
    \item \textbf{RQ2}: How important is each type of signal to the framework? 
    \item \textbf{RQ3}: How well does our framework work for student and TA models of different sizes? 
    \item \textbf{RQ4}: Does extending the framework to two-stage training boost more performance?
\end{itemize}

\noindent \textbf{Dataset} \quad
Since our work focuses on the scenario with sufficient unlabeled data but limited annotation budget, following the previous work \cite{wei2022chain, yao2022react}, we evaluate our framework on four datasets with a large training set: HotpotQA \cite{yang2018hotpotqa}, GSM8K \cite{cobbe2021training}, Aqua \cite{ling2017program} and CommonSenseQA \cite{talmor2019commonsenseqa}. We choose HotpotQA, the closed-book question and answering (QA) dataset, for evaluation with ReAct prompting \cite{yao2022react}. For GSM8K and Aqua, the arithmetic reasoning task and CommonSenseQA, the commonse sense reasoning task, we evaluate performance with CoT prompting \cite{wei2022chain}. In Table \ref{table:dataset}, we present the budget number, test number, prompting method, and evaluation metric of all four datasets.

\begin{table}[t]
\small
\centering
\resizebox{\columnwidth}{!}{%
\begin{tabular}{cccccc}
\hline
Dataset & \begin{tabular}[c]{@{}c@{}}\# Annotation \\ Budget\end{tabular} & \# Test & \begin{tabular}[c]{@{}c@{}}Prompting \\ Method\end{tabular} & Metric                                                      & Task Type                                                       \\ \hline
HotpotQA & 2,000                                                           & 1,000   & ReAct                                                       & Accuracy & \begin{tabular}[c]{@{}c@{}}Closed-book \\ QA\end{tabular}       \\ \hline
GSM8K    & 2,000                                                           & 1,319   & CoT                                                         & Accuracy                                                    & \begin{tabular}[c]{@{}c@{}}Arithmetic \\ reasoning\end{tabular} \\ \hline
Aqua     & 2,000                                                           & 254     & CoT                                                         & Accuracy                                                    & \begin{tabular}[c]{@{}c@{}}Arithmetic \\ reasoning\end{tabular} \\ \hline
CSQA     & 2,000                                                           & 1221     & CoT                                                         & Accuracy                                                    & \begin{tabular}[c]{@{}c@{}}Commonsense \\ reasoning\end{tabular} \\ \hline
\end{tabular}
}
\caption{\label{table:dataset} Dataset statistics.}
\end{table}
% \vspace{-2em}

% \vspace{0.1cm}
\noindent \textbf{Backbone models} \quad
For the teacher model, we use gpt-3.5-turbo based on InstructGPT 175B \cite{ouyang2022training} to generate the CoT or ReAct trajectories. We use Vicuna-13B and Vicuna-65B \cite{vicuna2023} as TA models to evaluate the answers of the teachers and students. For the student models, we choose between the GPT-J-6B and Vicuna-13B models \cite{gpt-j}.

\vspace{0.1cm}
\noindent \textbf{Proposed framework} \quad
We present two variants of the three-component framework with different signal weights as discussed in Section \ref{sec:annotation}: In our first framework, we use $\alpha=1$ and $\beta=0$ (\textbf{TA-finetune (I)}); In our second framework (\textbf{TA-finetune (T)}, we use $\alpha=\beta=1$ to compute annotation criteria. Both frameworks use the TA-teacher signal $s_e$ as Eq~\eqref{eq2}. Throughout the experiments sections, we use ``$\alpha = 1$'' and ``fine-tune with $s_i$'', ``$\beta = 1$'' and ``fine-tune with $s_t$'' interchangeably.

\vspace{0.1cm}
\noindent \textbf{Baseline models} \quad
We design several baseline experiments to show the superiority of our proposed methods. The first group of baseline methods is the few-shot ICL methods: \textbf{Student-ICL}, \textbf{TA-ICL}, and \textbf{Teacher-ICL} with the corresponding prompting methods, that is, ReAct for the HotpotQA dataset and CoT for the others \cite{wei2022chain, yao2022react}. The other baseline model is \textbf{Random-finetune}, the classical sequence-level KD pipeline for LLMs \cite{ho2022large, yao2022react, he2023teacherlm}, where the examples are sampled randomly until the budget is met, and all annotations from the teacher model are used for student fine-tuning.
% \textcolor{red}{
Although there are other sequence-level KD methods, some depend on ground-truth labels \cite{li2023turning, wang2023making, chen2023mcc, liu2023mind, an2023learning, zhao2023multistage}, not applicable in an unlabeled setting; those generate more augmented rationales \cite{liang2023mint, shridhar2023distilling, feng2023citing} or more augmented questions \cite{yu2023metamath}, which are not achievable with a fixed budget; and those incorporate additional loss \cite{hsieh2023distilling}, which is orthogonal to our KD method.
% }
%representing that we fine-tune the student model with annotations without TA evaluation and that the annotated samples are randomly sampled from the training set. 
% [yy: Can we just remove this paragraph? I didn't find it very informative..]Note that for all the proposed fine-tuning models and baseline models we fine-tune the student on the CoT or ReAct trajectories. 
% For vanilla fine-tuning (fine-tune on the final answer), since previous work has shown the superiority of fine-tuning on the CoT and ReAct rationales, we choose not to report the performance of vanilla fine-tuning \cite{ho2022large, yao2022react}.
% \citet{ho2022large} proposes the diverse reasoning to augment LM fine-tuning, where they generate multiple reasoning explanations for one question and fine-tune on various rationales. However, in our limited annotation budget setting, using the diverse reasoning leads to much fewer samples to annotate and fine-tune, so we do not regard Random-finetune with diverse reasoning as a baseline model.
The details of the model configuration are included in Appendix \ref{sec:implementation}.

%% file: 6_result.tex
\section{Results}
\label{sec:results}
\subsection{RQ1: Comparison with Baseline Methods}
\label{sec:rq1}
%\karamai{test}
\begin{table}[!t]
\centering
\resizebox{\linewidth}{!}{%
\begin{tabular}{lccccc}
\hline
Method          & HotpotQA       & GSM8K          & Aqua           & CSQA           & Avg.           \\ \hline
Random          & 0.00           & 0.00           & 20.00          & 20.00          & 10.00          \\ \hline
\multicolumn{6}{c}{\textbf{Teacher: GPT-3.5}}                                                        \\ \hline
Few-shot ICL    & 27.54          & 73.50          & 52.55          & 75.02          & 57.15          \\ \hline
\multicolumn{6}{c}{\textbf{TA: Vicuna-13B}}                                                          \\ \hline
Few-shot ICL    & 17.57          & 18.80          & 22.00          & 60.98          & 29.84          \\ \hline
\multicolumn{6}{c}{\textbf{Student: GPT-J-6B}}                                                       \\ \hline
Few-shot ICL    & 5.20           & 4.20           & 21.30          & 21.91          & 13.15          \\
Random-finetune & 13.10          & 18.57          & 13.93          & 57.57          & 25.78          \\
% TA-finetune     & 16.20          & 18.87          & 25.51          & 56.43          & 29.25          \\
TA-finetune (I) & 16.80          & \textbf{19.33} & 23.86          & \textbf{60.36} & \textbf{30.09} \\
TA-finetune (T) & \textbf{17.70} & 18.20          & \textbf{26.77} & 57.41          & 30.02          \\
% TA-finetune (S) & 15.90          & 19.00          & 22.60          & 57.41          & 28.73          \\ 
\hline
\end{tabular}
}
\caption{\label{table:results} \textbf{Performance of proposed three-component KD framework.} Accuracy (\%) of teacher (GPT-3.5), TA (Vicuna-13B), student (GPT-J-6B) models on four datasets. ``Random'' refers to random-guess performance in multiple-choice tasks. The results of student fine-tuned models with our framework can consistently outperform random fine-tuning without any signals and even better than the ICL of TA models.}
\end{table}
%\textbf{[YY: I think we need to rename the table and experiment findings if want to switch student and teacher signal to `signal for annotation` and `signal for filtering out low-quality annotation`]}

\noindent \textbf{Three-component KD framework outperforms Random-finetune and TA ICL} \quad
We use GPT-J-6B as the student model and Vicuna-13B as the TA model. We experiment with two variants of the proposed framework: TA-finetune (I) and TA-finetune (T). As shown in Table \ref{table:results}, our proposed framework shows around $16.7\%$ improvement compared to the Random-finetune baseline. It is also interesting to see that the fine-tuned GPT-J-6B model with the proposed framework can outperform the Vicuna-13B ICL model, which is already fine-tuned with 70k ChatGPT conversations \cite{vicuna2023}. 
 %shows the results of our proposed models and other baselines in different datasets. 
%We find that the ICL results continuously increase from student to TA to teacher model, but the performance of the teacher model in all tasks cannot exceed 80\%, which means that the proportion of noise in the teacher annotations cannot be easily ignored. 
%From Table \ref{table:results},  the proposed methods TA-finetune (I) and TA-finetune (T) with student and TA-Teacher signals can achieve the best two average performance among all tasks, and using disagreement (TA-finetune (I)) can increase the results of the Random-finetune from 25.78\% to 30.09\%, the relative improvement 16.71\%.

Note that the teacher ICL results are only $57.15\%$ on average, which means that the proportion of noise in the teacher annotations cannot be easily ignored. Another example in point is the multiple-choice QA Aqua dataset, where the fine-tuned student model without any signal (13.93\%) actually performs worse than its ICL counterpart (21.30\%) as well as the random guess baseline (20.00\%), and is primarily caused by the noise from teacher annotations. 
% The student's fine-tuning result without any signal (13.93\%) is worse than the student ICL (21.30\%) and the random guess (20.00\%), which are caused by the noise in teacher annotations.
An example of noisy teacher annotations in the Aqua dataset - the teacher model generates rationales with ``no answer'' as the final output when it cannot find a satisfactory result among the given choices. The student model imitates the teacher's behavior and generates the similar ``no answer'' outputs, leading to a worse result than ICL.
% \textbf{[YY: I didn't follow this] }
This case demonstrates the necessity of our annotation filtering signals to remove low-quality annotations. We also experiment with excluding the teacher annotations with a final ``no answer'' output and get 22.59\%, and our methods with signals still achieve better performance.

%Our two proposed TA-finetune frameworks can even outperform the average ICL accuracy of the TA model, which means that on a specific task, compared with the TA model, we can directly utilize the student model fine-tuned with signals for further inference to save the computation cost. 

\subsection{RQ2: Ablation Study of Signals}
\label{sec:rq2}
\begin{table}[!h]
\begin{subtable}{1\linewidth}
\centering
\resizebox{\linewidth}{!}{%
\begin{tabular}{ccccccc}
\hline
\multicolumn{2}{c}{Signal}                                                                                                          & \multirow{2}{*}{HotpotQA} & \multirow{2}{*}{GSM8K} & \multirow{2}{*}{Aqua} & \multirow{2}{*}{CSQA} & \multirow{2}{*}{Avg.} \\ \cline{1-2}
\textbf{$s_i$} & \textbf{$s_e$} &                           &                        &                       &                       &                       \\ \hline
N                                                                & N                                                                & 13.10                     & 18.57                  & 13.93                 & \textbf{57.57}        & 25.78                 \\
N                                                                & Y                                                                & \textbf{16.20}            & \textbf{18.87}         & \textbf{25.51}        & 56.43                 & \textbf{29.25}        \\ \hline\hline
Y                                                                & N                                                                & 13.80                     & \textbf{19.70}         & 15.91                 & 59.13                 & 27.14                 \\
Y                                                                & Y                                                                & \textbf{16.80}            & 19.33                  & \textbf{23.86}        & \textbf{60.36}        & \textbf{30.09}        \\ \hline
\end{tabular}
}
\caption{\label{table:rq2:conf_results} \textbf{Effects of TA-Teacher signal.} Results of fine-tuned students (GPT-J-6B) when considering our TA-Teacher signal. Signal $s_t$ is not used for experiments in this table.}
\end{subtable}

\begin{subtable}{1\linewidth}
\centering
\resizebox{\linewidth}{!}{%
\begin{tabular}{ccccccc}
\hline
\multicolumn{2}{c}{Signal}                                                                                                           & \multirow{2}{*}{HotpotQA} & \multirow{2}{*}{GSM8K} & \multirow{2}{*}{Aqua} & \multirow{2}{*}{CSQA} & \multirow{2}{*}{Avg.} \\ \cline{1-2}
\textbf{$s_i$} & \textbf{$s_t$} &                           &                        &                       &                       &                       \\ \hline
N                                                                & N                                                                 & 13.10                     & 18.57                  & 13.93                 & 57.57                 & 25.78                 \\
Y                                                                & N                                                                 & 13.80                     & 19.70                  & \textbf{15.91}        & 59.13                 & \textbf{27.14}        \\
N                                                                & Y                                                                 & \textbf{14.20}            & \textbf{20.30}         & 12.44                 & \textbf{59.71}        & 26.66                 \\ \hline
\end{tabular}
}
\caption{\label{table:rq2:student_results}  \textbf{Effects of signals for annotation.} Results of fine-tuned student models (GPT-J-6B) on four datasets when adding or not TA-Student signals or student internal signals. Signal $s_e$ is not used in the experiment of this table.}
\end{subtable}
\caption{Ablation study of all three signals. ``Y'' and ``N'' represent ``yes'' and ``no'' to indicate the signal use.}
\end{table}

%\textbf{[YY: can you rephrase this subsection to be aligned with the $\alpha$ and $\beta$ setup? Also need to change the column names in Table 4, 5 if necessary]}

After verifying the superiority of the overall framework on Random-finetune, we ask whether each component of signal presents its own functionality. 
In this section, we use the similar model and dataset setup in Section \ref{sec:rq1} and analyze the effects of each signal: TA-Teacher, TA-Student, and Student internal signal through ablation analysis. 
\vspace{0.1cm}

\noindent \textbf{TA-teacher signal brings salient performance boost} \quad
First, we compare the fine-tuned performance when adding or not the TA-Teacher signal in the condition of with or without the student internal signals $s_i$ and show the results in Table \ref{table:rq2:conf_results}.  

From Table \ref{table:rq2:conf_results}, we can observe that on average, excluding the uncertain annotations effectively boosts the performance, leading to the growth of the average accuracy from 25.78\% to 29.25\% or from 27.14\% to 30.09\% for fine-tuned with student disagreement ($s_i$) or not, respectively. Furthermore, the effects are more significant on more challenging datasets, such as HotpotQA and Aqua. 
This could be attributed to a significant improvement in data quality for these challenging datasets. We find that the ratio of the annotations with correct answer increases from 27.54\% to 49.80\% and from 52.55\% to 75.57\% for HotpotQA and Aqua datasets. 
% The possible reason for the comparable results of adding TA-Teacher signals on GSM8K and CommonSenseQA can be attributed to the lack of a large improvement in data quality in the fine-tuning dataset. 
% We find that when filtering by the TA-confidence, the ratio of teacher annotations with correct answers increases from 73.50\% to 80.64\% and from 75.02\% to 75.85\% for GSM8K and CommonSenseQA dataset, respectively. Considering the smaller data size after filtering (from 2,000 to 1,689 and from 2,000 to 1,509 for GSM8k and CommonSenseQA), smaller effects for TA-Teacher signals on these two datasets follow our expectation.
\vspace{0.1cm}

\noindent \textbf{Signals for annotation show their effectiveness} \quad
We separately demonstrate the effects of two types of signals for annotation (student internal $s_i$ and TA-student signal $s_t$) using only one of them to actively select the questions to annotate. We summarize the results in Table \ref{table:rq2:student_results}. The average results in Table \ref{table:rq2:student_results} show that the use of either of the signals for annotation can increase performance from 25.78\% to 27.14\% or 26.66\%, respectively, demonstrating the effectiveness of selecting uncertain samples to fine-tune the student model. 

From the ablation study, we find that all types of signal have their own functionalities and orthogonal to each other. The performance of using only one type of signal can still exceed the results of the Random-finetune framework, and the combination of signals will lead to a more performance boost.

\subsection{RQ3: Generalization on Students and TAs}
The ablation analysis presents the effectiveness of each signal, and we examine whether the effect of the framework can be generalized to student or TA models of different sizes. For student models, we additionally use Vicuna-13B as a student model and compare with the results in Table \ref{table:results} with GPT-J-6B as the student model. For TA models, we will show the effects with various TA models in two aspects: TA-teacher signal and TA-student signal.

\subsubsection{Generalizations on Students Models}

\begin{table}[!htb]
\centering
\resizebox{\linewidth}{!}{%
\begin{tabular}{ccccccc}
\hline
\multirow{2}{*}{Student model} & \multicolumn{2}{c}{Signal}                                                                                                           & \multirow{2}{*}{GSM8K} & \multirow{2}{*}{Aqua} & \multirow{2}{*}{CSQA} & \multirow{2}{*}{Avg.} \\ \cline{2-3}
                               & \textbf{$s_i$} & \textbf{$s_e$} &                        &                       &                       &                       \\ \hline
GPT-J-6B                       & Y                                                                & Y                                                                 & 19.33                  & 23.86                 & 60.36                 & 34.51                 \\ \hline
Vicuna-13B                     & N                                                                & N                                                                 & 38.06                  & 16.14                 & 71.41                 & 41.87                 \\
Vicuna-13B                     & Y                                                                & N                                                                 & 39.58                  & 19.37                 & \textbf{71.74}        & 43.56                 \\
Vicuna-13B                     & Y                                                                & Y                                                                 & \textbf{40.71}         & \textbf{29.92}        & 71.33                 & \textbf{47.32}        \\ \hline
\end{tabular}
}
\caption{\label{table:rq3:student-models} \textbf{Effects of student model scale.} Results of fine-tuned student models with multiple scales. We apply Vicuna-13B and GPT-3.5 as TA and teacher models.To avoid the inductive bias of self-refinement of Vicuna-13B, we do not use TA-student signal $s_t$.}
\end{table}

% [yy: for table 5, maybe also add TA model column or add it in the Table description so it's more clear?]

\noindent \textbf{Signals are effective for students of various sizes} \quad
We try Vicuna-13B as the student, and use another Vicuna-13B as the TA and GPT-3.5 as the teacher. We fine-tune Vicuna-13B as three signal combinations: no signals, only student internal signal $s_i$, and student internal $s_i$ + TA-teacher signals $s_e$. We show the results of GPT-J-6B and Vicuna-13B as the student model, respectively, in Table \ref{table:rq3:student-models}. Note that fine-tuning Vicuna-13B model on HotpotQA dataset will cause out-of-memory issues due to the large length of ReAct trajectories, so we only report the fine-tuned performance on other datasets.

We can observe that in Table \ref{table:rq3:student-models}, fine-tuning with the Vicuna-13B student model leads to much better performance than the GPT-J-6B model, indicating that a larger model size will cause better fine-tuning results, which is consistent with previous findings in \citet{ho2022large}. The best results for Vicuna-13B student models achieved with both TA-Student and TA-Teacher signals also verifies the generalizability of our framework on larger student models.

\subsubsection{Generalizations on TA Models}
TA models are used to examine both student and teacher answers. An intuitive question could be: How large could a model be an suitable choice of the TA? 
To explore this question, we use GPT-J-6B as the student, GPT-J-6B/Vicuna-13B/Vicuna-65B as the TA candidates, and GPT-3.5 as the teacher. To eliminate the other signals' influence, we experiment with only one signal: TA-teacher signal $s_e$ or TA-student signal $s_t$ in this part.

\begin{figure}[t]
     \centering
     \includegraphics[width=\linewidth]{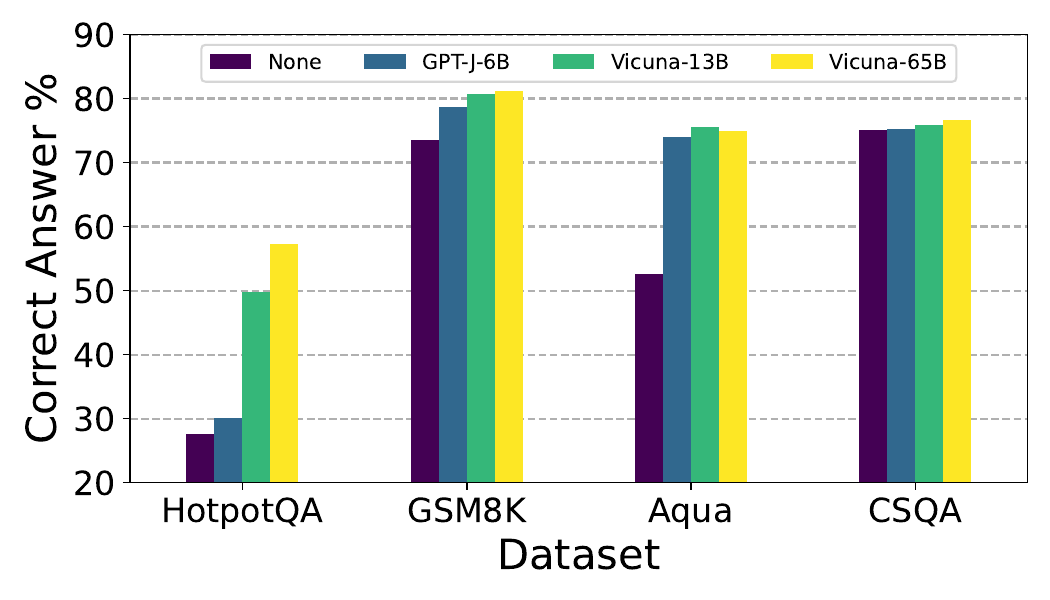}
     \caption{\textbf{Data quality of the fine-tuning bucket}. The proportion (\%) of samples with correct annotations in the fine-tuning bucket when utilizing different scales of TA models in the TA-Teacher signal $s_e$. }
     \label{fig:rq3:ta-acc}
\end{figure}

\noindent \textbf{Choice of TA-Teacher signals} \quad
Since the TA model in TA-teacher signal $s_e$ filters incorrect teacher annotations to improve data quality in the fine-tuning bucket $B_f$, our hypothesis is that the data quality in $B_f$ scales with the size of the TA, which influences the student fine-tuning result. 
% To verify the hypothesis, we first examine the data quality in the bucket $B_f$ with different TA models by checking the ratio of questions with the correct annotations. 
To verify the hypothesis, 
% \textcolor{red}{
we first collect the final teacher answer (the single answer) in different fine-tuning buckets. Then, we calculate the accuracy of the answer collection for each bucket, using the ground-truth answer from the original labeled dataset.
% }
We visualize the proportion of questions with correct teacher annotations with different TAs in Figure \ref{fig:rq3:ta-acc}.
% \begin{table}[!htb]
% \centering
% \resizebox{\linewidth}{!}{%
% \begin{tabular}{cccccc}
% \hline
% TA model   & HotpotQA       & GSM8K          & Aqua           & CSQA           & Avg.           \\ \hline
% None       & 27.54          & 73.50          & 52.55          & 75.02          & 57.15          \\
% GPT-J-6B   & 30.04          & 78.71          & 74.00          & 75.18          & 64.48          \\
% Vicuna-13B & 49.80          & 80.64          & \textbf{75.57} & 75.85          & 70.30          \\
% Vicuna-65B & \textbf{57.28} & \textbf{81.25} & 75.00          & \textbf{76.59} & \textbf{72.53} \\ \hline
% \end{tabular}
% }
% \caption{\label{table:rq3:ta-acc} \textbf{Data quality of the fine-tuning bucket}. The proportion of samples with correct annotations in the fine-tuning bucket when utilizing different scales of TA models in the TA-Teacher signal. }
% \end{table}
Our observation is that with larger TAs, the fine-tuning data quality increases, but the improvement for model size from Vicuna 13B to 65B is smaller than that from GPT-J-6B to Vicuna-13B. 

Then, we fine-tune the student model GPT-J-6B with the bucket $B_f$ evaluated by different TA models. The results can be found in Table \ref{table:rq3:ta-results}. From Table \ref{table:rq3:ta-results}, we obtain the highest fine-tuning accuracy with the best TA, but the marginal improvement in data quality from 13B to 65B revealed in Figure \ref{fig:rq3:ta-acc} leads to comparable performance. 

\begin{table}[!htb]
\centering
\resizebox{\linewidth}{!}{%
\begin{tabular}{cccccc}
\hline
TA model   & HotpotQA       & GSM8K          & Aqua           & CSQA           & Avg.           \\ \hline
GPT-J-6B   & 13.20          & 17.43          & 23.15          & 52.74          & 26.63          \\
Vicuna-13B & 16.20          & 18.87          & \textbf{25.51} & 56.43          & 29.25          \\
Vicuna-65B & \textbf{17.12} & \textbf{20.47} & 24.41          & \textbf{57.98} & \textbf{30.00} \\ \hline
\end{tabular}
}
\caption{\label{table:rq3:ta-results}   \textbf{Effects of TA model scale in the TA-Teacher signals.} Results of the fine-tuned student (GPT-J-6B) with different sizes of TAs in the TA-Teacher signals.}
\end{table}

\noindent \textbf{Choice of TA-student signals} \quad
We vary the size of TA in the TA-student signals $s_t$ to characterize the uncertainty of the student's inference. We use Vicuna-13B or Vicuna-65B as the TA to evaluate the student's inference from the GPT-J-6B model. 
% Here, we do not use TA-Teacher signals $s_e$ to filter teacher annotations to eliminate confounding variables. 
The fine-tuning results of the student model with only the TA-student signal are shown in Table \ref{table:rq3:ta-student-results}. 

\begin{table}[!htb]
\centering
\resizebox{\linewidth}{!}{%
\begin{tabular}{cccccc}
\hline
TA model & HotpotQA & GSM8K & Aqua & CSQA & Avg. \\ \hline
Vicuna-13B & \textbf{14.20}            & 20.30                  & \textbf{12.44}        & 59.71                 & \textbf{26.66}        \\
Vicuna-65B & 13.40                     & \textbf{20.47}         & 12.20                 & \textbf{59.78}        & 26.46                 \\ \hline
\end{tabular}
}
\caption{\label{table:rq3:ta-student-results}  \textbf{Effects of TA model scale in the TA-student signals.} Results of the fine-tuned student (GPT-J-6B) with two different sizes of TAs in the TA-student signals.}
\end{table}

From comparison with different TA models in the TA-student signal from Table \ref{table:rq3:ta-student-results}, we can obtain comparable results when using a more advanced TA model to characterize the uncertainty of student’s generations. We conjecture that the difference in difficulty between the annotation buckets $B_a$ of different TA models, especially from 13B to 65B, is not that large, which may not show much significant effect on the final fine-tuning results. 

Results in Tables \ref{table:rq3:ta-results} and \ref{table:rq3:ta-student-results} indicate a trade-off between the TA model sizes and the fine-tuning performance. 
Using a more advanced TA can improve some performance. However, in some scenarios, the computation resource can not afford a large TA model, such as the 65B model, then using the 13B model, which can effectively improve the fine-tuning data quality, is also a reasonable choice.

Moreover, we use different budget numbers to fine-tune the student model and present the detailed results in Appendix \ref{sec:different_budgets}. The observation from model performance can verify the generalization of our framework with various budget numbers.

\subsection{RQ4: Effects of Two-stage Training}
\label{sec:rq4}

\begin{table}[!htb]
\centering
\resizebox{\linewidth}{!}{%
\begin{tabular}{cccccccc}
\hline
\multicolumn{1}{l}{\multirow{2}{*}{Stage}} & \multicolumn{2}{c}{Signal}                                                                                                           & \multirow{2}{*}{HotpotQA} & \multirow{2}{*}{GSM8K} & \multirow{2}{*}{Aqua} & \multirow{2}{*}{CSQA} & \multirow{2}{*}{Avg.} \\ \cline{2-3}
\multicolumn{1}{l}{}                              & \textbf{$s_i$} & \textbf{$s_e$} &                           &                        &                       &                       &                       \\ \hline
1                                         & Y                                                                & N                                                                 & \textbf{13.80}            & 19.70                  & 21.26                 & \textbf{59.13}        & 28.47                 \\
2                                         & Y                                                                & N                                                                 & 13.50                     & \textbf{20.17}         & \textbf{23.46}        & 58.86                 & \textbf{29.00}        \\ \hline
1                                         & Y                                                                & Y                                                                 & 16.80                     & 19.33                  & 23.86                 & 60.36                 & 30.09                 \\
2                                         & Y                                                                & Y                                                                 & \textbf{16.90}            & \textbf{21.30}         & \textbf{25.98}        & \textbf{60.36}        & \textbf{31.14}        \\ \hline
\end{tabular}
}
\caption{\label{table:rq4:two-stage} \textbf{Effects of two-stage extension}. Results of fine-tuned student models (GPT-J-6B) on four datasets when extending from one-stage to two-stage training. Signal $s_t$ is not used in the experiment of this table.}
\end{table}

As we described in Section \ref{sec:extension}, we extend our one-stage training to two-stage training to better utilize the student internal signals. For the model and dataset setup, we use the same setting as in Section \ref{sec:rq1}, which means that for the two-stage training, we use 200 examples for the first warm-up stage and the remaining 1,800 budgets for the second stage. For signal setup, we choose the student internal signal $s_i$ and experiment under two conditions: with or without TA-Teacher signals $s_e$. We summarize the results in Table \ref{table:rq4:two-stage}.

\vspace{0.5em}
\noindent \textbf{Two-stage extension obtain better results in most cases} \quad
From the results in Table \ref{table:rq4:two-stage}, we observe that extending the one-stage to the two-stage framework obtains a relative improvement of 1.86\% and 3.48\% for fine-tuning with or without TA-Teacher signals, respectively, and compared with the Random-finetune baseline, the two-stage framework brings the highest relative improvement of \textbf{20.79\%}. Our extension obtains better results in 6/8 comparison cases, which verifies the effectiveness of the extension. The improvement can be attributed to more reliable signals $s_i$ extracted by the student model in the second stage. We show the intermediate results in Appendix \ref{sec:two_stage_details} to interpret the improvement of the two-stage training. Moreover, we extend our fine-tuning to more stages to a curriculum learning pipeline, and the results of four-stage learning are presented in Appendix \ref{sec:cl}. 

%% file: 7_conclusion.tex
\section{Conclusions}
In this paper, we propose a novel unified framework to resolve two challenges in the current teacher-student knowledge distillation process. Our framework utilizes a third TA model and signals from the student and teacher side to actively select samples and improve data quality in teacher annotations. Additionally, to better utilize student signals, we also extend our framework to two-stage training. With a limited budget for teacher annotations on an unlabeled dataset, our extensive empirical evaluations show that the proposed framework can significantly increase KD results with signals and all proposed signals show its own effectiveness.

\section{Limitations}
In our framework, we apply several types of signals: student intenal signal, TA-student and TA-teacher signal to refine the knowledge distillation of LLM.
% For signals for annotation $s_i$ and $s_t$, we use $\alpha$, $\beta$, the signal weights and $C_1$, $C_2$, the complexity sets to control the signal usage and the annotation criteria. Since our aim of using $s_i$ and $s_t$ is to choose examples that are not too easy or too hard for the student to learn, we set $C_1=\{2,3\}$ and $C_2=$ \{confident, not confident\}. Our work shows the effects of our method with different values of $\alpha$ and $\beta$ (Section \ref{sec:rq1} and \ref{sec:rq2}) but does not test the sensitivity of the framework to the choice of these hyperparameters in $C_1$, $C_2$ and $n$. In future work, we can also explore a larger value of $n$ and more confidence levels for the TA model to classify teacher or student annotations.

In addition to these internal and external signals, we can try other signals developed in existing works about active learning \cite{zhang2022survey, settles2009active, li-etal-2024-improving}. For example, we can aggregate the model confidence on each generated token in the rationales as the student or teacher's confidence on their annotations. Moreover, in addition to querying the TA model of confidence on the final answers, we can also ask the TA to evaluate the intermediate reasoning steps of the teacher or student models on each question, which could be another type of external signal. 

% \textcolor{red}{
We have verified the effectiveness of our framework on multiple models, prompting methods, and datasets. In related work, we have disccused multiple sequence-level KD methods, using more advanced fine-tuning loss \cite{hsieh2023distilling} and more complex generated rationales \cite{shridhar2023distilling} to improve KD performance. These methods are complementary to our three-component framework, aiming to increase the results with sufficiently large teacher-annotated data. Our method could further enhance their efficiency in fixed-budget and unlabeled data settings. We leave the exploration of the combination with existing works to the future.
% }

The student models (GPT-J-6B and Vicuna-13B) in the experiments are fine-tuned on 8 NVIDIA A100 GPUs, which may not be accessible to everyone and have a negative environmental impact. For further experiments, we can extend our experiments to smaller language models with fewer than 1 billion parameters as the student model. Furthermore, we can also use more advanced teacher models, such as GPT-4 \cite{openai2023gpt4}, to observe whether a better teacher model leads to a greater improvement in performance.

% \section*{Acknowledgments}

% The acknowledgments should go immediately before the references. Do not number the acknowledgments section.
% \textbf{Do not include this section when submitting your paper for review.}

\section*{Acknowledgments}
We would like to thank  the anonymous reviewers as well
as Paiheng Xu for reviewing the article and for providing helpful comments and suggestions. 

%% file: 8_appendix.tex
\section{Implementation Details}
\label{sec:implementation}
\subsection{Signal and annotation generation}
For all inference trajectories of the student models for student internal signal, TA-student signal and test inference, following the previous work, we set the maximum sequence length to 1,024 \cite{ho2022large, kojima2022large}. For teacher models' annotations, we set the maximum sequence length to 2,048 for GPT-3.5. 

We utilize greedy search in decoding for all generations, except for the students' generations for the collection of student internal signals $s_i$, where we use stochastic temperature sampling with the same temperature value 0.7 as in the previous work \cite{diao2023active, ho2022large, wang2022self, zhou2024explore, zhou2024calibrated, liu2024large}.

We use the same few-shot ICL prompts as in previous work to generate student or teacher annotations for different datasets \cite{wei2022chain, wang2024enhancing, diao2023active, wang2024mementos} and for the crafted confidence prompt $P_c$, the details and exemplars inside the prompts are shown in Tables \ref{table:ta_conf_prompt} and \ref{table:examples_ta_conf}. 

We call the gpt-3.5-turbo function from OpenAI to generate teacher annotations and rationales. The price of this API is \$0.0015 / 1K tokens for inputs and \$0.002 / 1K tokens for output. The total expenditure on API usage is \$ 207.05, including preliminary exploration.

\subsection{Student model fine-tuning}
For the fine-tuning of the student model, we base our implementation on the Pytorch\footnote{\url{https://pytorch.org/}}, Huggingface transformer\footnote{\url{https://huggingface.co/}}, and Stanford Alpaca\footnote{\url{https://github.com/tatsu-lab/stanford_alpaca}}. We use AdamW as our optimizer with a learning rate of $2\mathrm{e}{-5}$ and a weight decay of 0.01 with linear scheduler, batch size of 1, and trained for 5 epochs with the early stopping mechanism.

% \begin{algorithm}[t!]
% \caption{\textbf{Three-component Knowledge Distillation Framework}}
% \label{alg:ta}
% \begin{algorithmic}[1]

% \State \textbf{Input:} Unlabeled dataset $U$, student model $M_{s}$, TA model $M_{t}$, teacher model $M_{e}$, confidence prompt $P_{c}$, annotation criteria annotate$(\cdot)$, confidence set $C$, annotation bucket $B_{a}$, fine-tuning bucket $B_{f}$
% \State \textbf{Output:} The desired fine-tuning bucket $B_{f}$

% % \vspace{0.15cm}
% % \Statex \textbf{\sf \textcolor{gray}{/* STEP 1. Check if input is worth annotating by collecting the student signals */}}
% \For{each input $x$ from dataset $U$}
%         \State $t_{i}(x) =M_{s}(x, P_i) \text{ for $i\in\{1,\cdots, n\}$ }$
%         \State $s_{i} (x)$ = $\text{Uniq}(t_{1}(x) \dots t_{n}(x))$
%         \State $s_{t} (x)$ = $M_{t}({x, t_1, P_c})$
%         \If{$\text{annotate}(s_{i}, s_{t})$ is True} 
%             \State Add $x$ into annotate bucket $B_{a}$
%         \EndIf
% \EndFor

% % \Statex \textbf{\sf \textcolor{gray}{/* STEP 2. Ask teacher to annotate and use TA to filter the unconfident ones*/}}
% \For{each input $x$ from dataset $B_{a}$}
%         \State $t_{e}(x)$ = $M_{e}(x, P_{i})$
%         \State $s_{e}(x)$ = $M_{t}(x, t_e, P_{c})$
%         \If{$s_{e}$ in $C$} 
%             \State Add $x$ into fine-tuning bucket $B_{f}$
%         \EndIf
% \EndFor

% \State \Return{$B_{f}$}
% \end{algorithmic}
% \end{algorithm}

\section{Supplementary Experimental Results}

In this section, we will show the supplementary experimental results, including our framework with different budgets and the performance of extending our framework to curriculum learning.

\subsection{Generalizations on Different Budgets}
\label{sec:different_budgets}
In Section \ref{sec:rq1}, we have verified the effectiveness of our framework with 2,000 as the budget number, and to show the generalization of our framework on different budget numbers, we will repeat the experiments with the same model and dataset setup with the budget numbers 1,000 and 3,000, respectively. We present the results in Table \ref{table:ap1_main}.

\begin{table}[!htb]
\centering
\resizebox{\linewidth}{!}{%
\begin{tabular}{lccccc}
\hline
\multirow{2}{*}{Method} & \multirow{2}{*}{HotpotQA} & \multirow{2}{*}{GSM8K} & \multirow{2}{*}{Aqua} & \multirow{2}{*}{CSQA} & \multirow{2}{*}{Avg.} \\
                        &                           &                        &                       &                       &                       \\ \hline
\multicolumn{6}{c}{Budget: 1000}                                                                                                                     \\ \hline
Random-finetune         & 10.20                     & \textbf{15.46}         & 10.08                 & 53.97                 & 22.43                 \\
TA-finetune (I)         & \textbf{15.90}            & 13.94                  & 20.16                 & \textbf{59.46}        & \textbf{27.37}        \\
TA-finetune (T)         & 13.70                     & 13.12                  & \textbf{23.54}        & 57.41                 & 26.94                 \\ \hline
\multicolumn{6}{c}{Budget: 3000}                                                                                                                     \\ \hline
Random-finetune         & 13.40                     & 21.61                  & 13.93                 & 57.57                 & 26.63                 \\
TA-finetune (I)         & \textbf{17.10}            & 21.30                  & 21.02                 & \textbf{59.46}        & 29.72                 \\
TA-finetune (T)         & 16.20                     & \textbf{22.51}         & \textbf{24.17}        & 57.75                 & \textbf{30.16}        \\ \hline
\end{tabular}
}
\caption{\label{table:ap1_main} \textbf{Effects of budget number.} Results of fine-tuned student models (GPT-J-6B) on four datasets when setting the budget number to 1,000 or 3,000 respectively.}
\end{table}

From Tables \ref{table:ap1_main} and \ref{table:results}, we find that the performance of all methods improves a lot when the budget number increases from 1,000 to 2,000 and will get a comparable performance when the budget increases from 2,000 to 3,000. Our proposed three-component KD framework can outperform the random-finetune method in all three cases, which suggests that our framework's effectiveness can be generalized to different numbers of budgets. 

\subsection{Details of Two-stage Training}
\label{sec:two_stage_details}
For the interpretation of the improvement in two-stage training in Section \ref{sec:rq4}, we conjecture that it could be attributed to two reasons. First, after the warm-up stage, the better student model is more expert in the questions, so the student's internal signal $s_i$ should be more reliable.
Second, samples extracted by student internal signals $s_i$ (the uncertain questions for the student) should be more challenging in the second stage and continually fine-tuning the student with increasingly difficult data, leading to a deeper learning curve and a higher performance boost. To verify these two hypotheses, we present the accuracy of the student model and the difficulty level of the samples in the annotation bucket $B_a$ before and after the warm-up stage.

\begin{figure}[!ht]
     \centering
     \includegraphics[width=\linewidth]{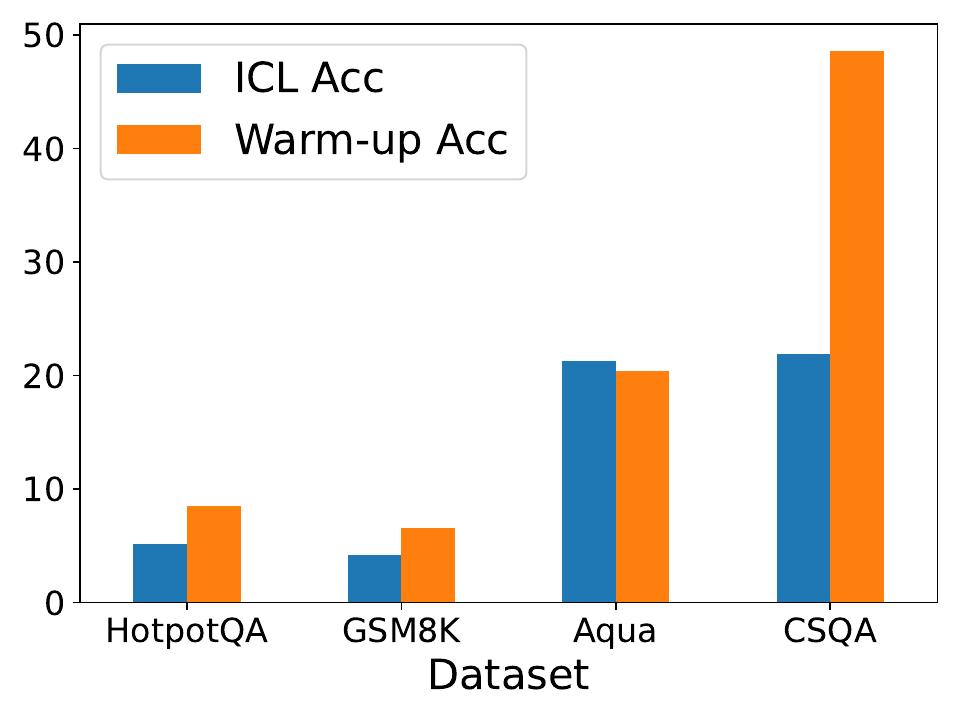}
     \caption{\textbf{Accuracy comparison of the student in the first and second stage.} The student ICL accuracy (ICL Acc) in the first stage and the student accuracy in the second stage (Warm-up Acc). After fine-tuning on a small dataset to warm up, the student model becomes more promising and expert in answering complex questions in most cases.}
     \label{fig:icl_acc}
\end{figure}

\begin{figure}[!ht]
     \centering
     \includegraphics[width=\linewidth]{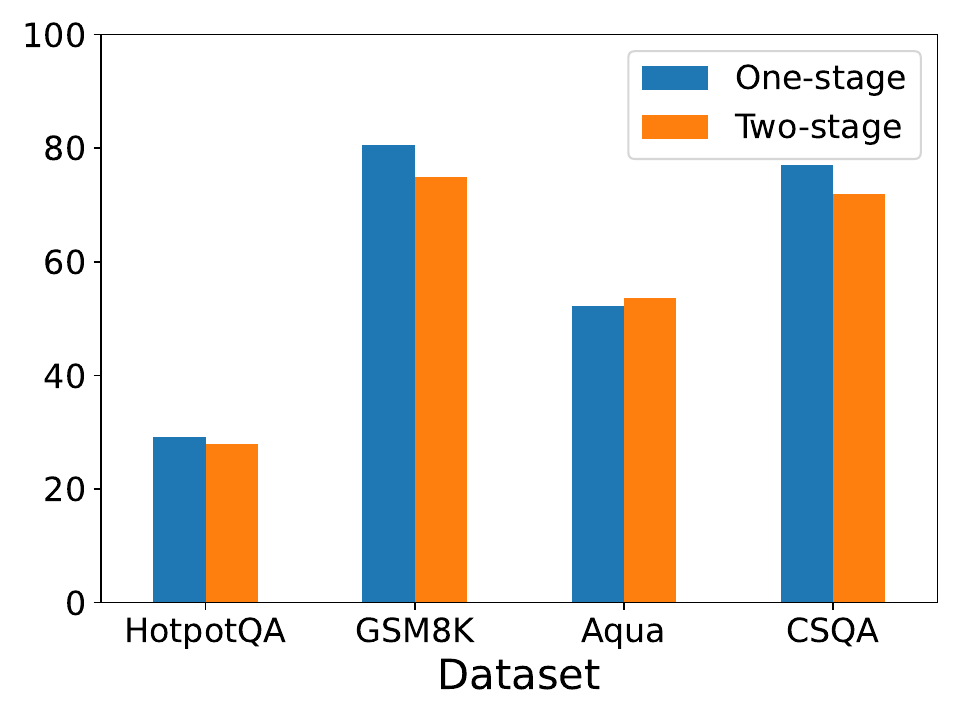}
     \caption{\textbf{Data difficulty in the annotation bucket.} Teacher model's accuracy on the annotation bucket $B_a$ in the first stage and second stage, respectively. The teacher model's accuracy on the questions in $B_a$ for the second stage is lower than that for the first stage and suggests more difficult and challenging questions in $B_a$.}
     \label{fig:hardness}
 \end{figure}

We first present the student ICL accuracy (ICL Acc) in the first stage and the student accuracy in the second stage (Warm-up Acc) in Figure \ref{fig:icl_acc}. We find that after fine-tuning on a small dataset to warm up, the student model becomes more promising and expert in answering complex questions in most cases. 

Then, we also quantify the difficulty of samples in the annotation bucket $B_a$ of in the first stage and second stage, respectively, by measuring the teacher model accuracy on questions in $B_a$. We visualize the accuracy of the teacher model of the samples in the annotation bucket in two stages in Figure \ref{fig:hardness}. We find that in most datasets, the teacher model's accuracy on the questions in the annotation bucket for the second stage is lower than that for the first stage and suggests more difficult and challenging questions in $B_a$.
This observation indicates that continually fine-tuning the student model on the slightly more difficult data causes a deeper learning curve, leading to a better performance. 

\subsection{Explorations on Curriculum Learning}
\label{sec:cl}
Inspired by the results of the two-stage training, our next question is whether the performance improvement will continue to increase when using more stages. We can frame multiple-stage training into the curriculum learning scope, which means that at each stage, we will have better student models and the samples extracted by student signals should be more difficult and challenging \cite{xu2020curriculum}. For our experiment setup, we divide our training into four stages and use 200, 400, 600, and 800 budgets (a total of 2,000) in each stage. For the final step, we will integrate all the annotations in every stage and continue fine-tuning the student model for a few epochs. We present the results of one-stage training and curriculum learning in Table \ref{table:cl}.

\begin{table}[!htb]
\centering
\resizebox{\linewidth}{!}{%
\begin{tabular}{ccccccc}
\hline
\multicolumn{1}{l}{\multirow{2}{*}{Training Method}} & \multicolumn{2}{c}{Signal}                                                                                                           & \multirow{2}{*}{HotpotQA} & \multirow{2}{*}{GSM8K} & \multirow{2}{*}{Aqua} & \multirow{2}{*}{Avg.} \\ \cline{2-3}
\multicolumn{1}{l}{}                              & \textbf{$s_i$} & \textbf{$s_t$} &                           &                        &                       &                       \\ \hline
One                                         & Y                                                                & N                                                                 & \textbf{13.80}            & \textbf{19.70}         & \textbf{15.91}        & \textbf{16.47}        \\
CL                                                & Y                                                                & N                                                                 & 11.21                     & 19.60                  & 14.96                 & 15.26                 \\ \hline
One                                         & Y                                                                & Y                                                                 & \textbf{16.80}            & \textbf{19.33}         & \textbf{23.86}        & \textbf{19.99}        \\
CL                                                & Y                                                                & Y                                                                 & 16.60                     & 18.57                  & 22.52                 & 19.23                 \\ \hline
\end{tabular}
}
\caption{\label{table:cl} \textbf{Effects of curriculum learning}. Results of fine-tuned student models (GPT-J-6B) on three datasets when extending the one-stage training to a curriculum learning method (four-stage training).}
\end{table}

From Table \ref{table:cl}, we do not observe improvement as in the results of the two-stage training, and we conjecture that there should exist multiple reasons to influence the results of curriculum learning, such as the learning rate for each stage and the number of budgets in each stage. Here, we keep the learning rate for each stage as the same value, which may cause the fine-tuning to a sub-optimal point. We leave the further exploration of curriculum learning of our framework as a future direction.

\subsection{Preliminary Explorations of Annotation Criteria}
\label{sec:pre_annotation}
% \textcolor{red}{
In Section \ref{sec:annotation}, we choose $C_1=\{2, 3\}$ as the annotation criteria of the internal signal $s_i$, and this hyperparameter choice is decided by our preliminary experiment. We compare the fine-tuning result with only $s_i$ (without signal $s_t$ and $s_e$) by choosing $C_1=\{2, 3\}$ and $C_1'=\{4, 5\}$ and without $s_i$ (randomly sampled) by fixing budget number 300 in the HotpotQA dataset. The model configuration is the same as in Section \ref{sec:rq1}. The result is shown in Table \ref{tab:pre_annotation}.
% }

\begin{table}[!htb]
\centering
\resizebox{\linewidth}{!}{%
\begin{tabular}{l|c|c|c|c}
\hline
Task     & Few-shot ICL  & Finetuned without $s_i$ & Finetuned with $C_1'$ & Finetuned with $C_1$ \\ \hline
HotpotQA & 5.20\%       & 8.50\%           & 7.50\%             & \textbf{10.50\%}                 \\ \hline
\end{tabular}
}
\caption{\textbf{Preliminary exploration of annotation criteria on HotpotQA dataset.} Results of the fine-tuned student model (GPT-J-6B) on HotpotQA when fine-tuned without $s_i$ (randomly sampled questions), with $s_i$ and $C_1=\{2, 3\}$ and $C_1'=\{4, 5\}$.}
\label{tab:pre_annotation}
\end{table}

% \textcolor{red}{
From Table \ref{tab:pre_annotation}, we find that the student model fine-tuned with $C_1=\{2, 3\}$ achieves the best performance, indicating that fine-tuning with $C_1=\{2, 3\}$ can bring the deepest learning curve, so we choose this hyperparameter choice in the following experiments. Furthermore, we observe that the model fine-tuned with the annotation criteria $C_1'$ cannot outperform random selection without signal $s_i$, suggesting that it is hard for the student to learn useful information from hard questions. 
% }

\begin{table*}[ht]
    \begin{subtable}{1\textwidth}
    \centering
    \begin{tabular}{m{40em}}
        \hline
        You are an advanced reasoning agent. You will be given a previous reasoning trial in which you were given access to a Docstore API environment and the answer to the question. Report the confidence about the answer from choices (a) very confident (b) confident (c) not confident (d) wrong answer. \\
        
        Here are some examples:
        \textcolor{red}{\{demonstrations\}}
        \\
        Previous Trial:
        \textcolor{red}{\{model generations\}} \\
        Confidence Choice:
        \\
        \hline
        
    \end{tabular}
    \caption{Prompt $P_c$ for HotpotQA dataset.}
    \end{subtable}
    \bigskip
    \begin{subtable}{1\textwidth}
    \centering
    \begin{tabular}{m{40em}}
        \hline
        You are an advanced \textcolor{red}{\{question type\}} agent. You will be given a previous \textcolor{red}{\{question type\}} question, the reasoning trail and the answer to the question. Report the confidence about the answer from choices (a) very confident (b) confident (c) not confident (d) wrong answer. \\
        
        Here are some examples:
        \textcolor{red}{\{demonstrations\}}
        \\
        Previous Trial:
        \textcolor{red}{\{model generations\}} \\
        Confidence Choice:
        \\
        \hline
        
    \end{tabular}
    \caption{Prompt $P_c$ for GSM8K, Aqua and CommonsenseQA dataset. For GSM8K and Aqua, \textcolor{red}{\{question type\}} is ``arithmetic reasoning'' and for CommonsenseQA, it is ``commonsense reasoning''.}
    \end{subtable}
    \caption{\label{table:ta_conf_prompt} Prompt $P_c$ for all datasets. \textcolor{red}{\{demonstrations\}} is a placeholder to add four demonstrations with provided ground-truth TA confidence for each dataset. \textcolor{red}{\{model generations\}} is the place to insert the teacher or student generated annotations.}
\end{table*}

\begin{table*}[ht]
    \begin{subtable}{1\textwidth}
    \centering
    \begin{tabular}{m{40em}}
        \hline
        Q: The Fiery Furnaces and Paul Westerberg, are of which nationality? \\
        Thought 1: I need to search The Fiery Furnaces and Paul Westerberg, find their nationality, and answer the question. \\
        Action 1: Search[The Fiery Furnaces] \\ 
        Observation 1: [truncated wikipedia excerpt] \\
        Thought 2: The Fiery Furnaces is an American band. I need to search Paul Westerberg next and find his nationality. \\
        Action 2: Search[Paul Westerberg] \\
        Observation 2: Paul Harold Westerberg (born December 31, 1959) is an American musician, best known as the lead singer, guitarist, and songwriter for The Replacements.[2] Following the breakup of The Replacements, Westerberg launched a solo career that saw him release three albums on two major record labels.. Following the release of his third solo album, Westerberg has been mostly releasing music that he has self-produced and recorded in his basement home studio.[3] He has also released two albums and an EP under the pseudonym Grandpaboy. In 2017, Westerberg released songs on SoundCloud as User 964848511 and on Bandcamp as Dry Wood Garage.[4][5][6]. In the late 1970s, Westerberg was working as a janitor for U.S. Senator David Durenberger,[7] and one day while walking home from work, he heard a band practicing Yes's "Roundabout" in a basement. \\
        Thought 3: Paul Westerberg is also an American musician. So The Fiery Furnaces and Paul Westerberg are both American.  \\
        Action 3: Finish[American] \\
        Confidence Choice: (a) very confident
        \\
        \hline   
    \end{tabular}
    \caption{One of the examples in the demonstrations of the Prompt $P_c$ for TA confidence generation for HotpotQA dataset}
    \end{subtable}
    \bigskip
    \begin{subtable}{1\textwidth}
    \centering
    \begin{tabular}{m{40em}}
        \hline
        Q: Abraham owns 80 square meters of unused land. He sold half of the land for \$50, and after a month, he sold another 1/4 of his land for \$30. He then sold the remaining land for \$3 per square meter. How much money will he be able to earn after selling all his unused land?
        \\
        A: Let's think step by step. Abraham owns 80 square meters of unused land. He sold half of the land for \$50. After a month, he sold another 1/4 of his land for \$30. He then sold the remaining land for \$3 per square meter. 80 - 30 = 50. Then he sold the remaining land for \$3 per square meter. 50 - 3 = 47. The answer is 47.
        \\
        Confidence Choice: (d) wrong answer
        \\
        \hline   
    \end{tabular}
    \caption{One of the examples in the demonstrations of the Prompt $P_c$ for TA confidence generation for GSM8K dataset}
    \end{subtable}
    \bigskip
    \begin{subtable}{1\textwidth}
    \centering
    \begin{tabular}{m{40em}}
        \hline
        Q: Alfred buys an old scooter for \$4700 and spends \$800 on its repairs. If he sells the scooter for \$5800, his gain percent is? Answer Choices: (A) 6.23\% (B) 5.45\% (C) 7\% (D) 8.12\% (E) 10\%
        \\
        A: Let's think step by step. The total cost of the scooter for Alfred is \$4700 + \$800 = \$5500. His gain is the selling price minus the cost price, which is \$5800 - \$5500 = \$300. The gain percent is (gain / cost price) * 100\% = (\$300 / \$5500) * 100\% = 5.45\% (rounded to two decimal places). The answer is B.
        \\
        Confidence Choice: (a) very confident
        \\
        \hline   
    \end{tabular}
    \caption{One of the examples in the demonstrations of the Prompt $P_c$ for TA confidence generation for Aqua dataset}
    \end{subtable}
    \bigskip
    \begin{subtable}{1\textwidth}
    \centering
    \begin{tabular}{m{40em}}
        \hline
        Q: When old conservative people observe a changing society, what do the experience? Answer Choices: (A) depression (B) resentment (C) hurt (D) anger (E) acceptance \\
        A: Let's think step by step. The answer should be the feeling that old conservative people experience when observing a changing society. Of the above choices, the closest feeling is resentment. The answer is B. \\
        Confidence Choice: (c) not confident
        \\
        \hline   
    \end{tabular}
    \caption{One of the examples in the demonstrations of the Prompt $P_c$ for TA confidence generation for CommonsenseQA dataset}
    \end{subtable}
    \caption{\label{table:examples_ta_conf} Examples in the demonstrations of the Prompt $P_c$ for TA confidence (TA-student signal $s_t$ and TA-teacher signal $s_e$) generation for all four datasets}
\end{table*}

\begin{table*}[ht]
    \begin{subtable}{1\textwidth}
    \centering
        \begin{tabular}{m{6em}|m{33em}}
            \hline
            \textbf{Question} & On Friday, Markeesha sold 30 boxes of crackers for her scout troop's fundraiser. On Saturday, she sold twice as many as on Friday. On Sunday, she sold 15 fewer than Saturday. How many boxes did she sell over the three days? \\ \hline
            \textbf{Generation 1} & Let's think step by step. Markeesha started with 30 boxes of crackers. On Friday, she sold 30 boxes. On Saturday, she sold 70 boxes. On Sunday, she sold 20 less than Saturday. So she sold 60 boxes - 20 = 40 boxes. The answer is 40. \\ \hline
            \textbf{Generation 2} & Let's think step by step. Markeesha started with 30 boxes. On friday, she sold 30 boxes. On saturday, she sold 60 boxes. On saturday, she sold 60 boxes - 15 = 45. The answer is 45. \\ \hline
            \textbf{Generation 3} & Let's think step by step. Markeesha started with 30 boxes. On Friday, she sold 30 boxes for her troop's fundraiser. On Saturday, she sold 60 boxes for her troop's fundraiser. On Sunday, she sold 15 fewer than Saturday. So she sold 30 + 60 - 15 = 45 boxes. The answer is 45. \\ \hline
            \textbf{Generation 4} & Let's think step by step. On Friday, she sold 30 boxes. On Saturday, she sold twice as many. So she sold 60 boxes. On Sunday, she sold 15 fewer than Saturday. So she sold 45 boxes. The answer is 45. \\ \hline
            \textbf{Generation 5} & Let's think step by step. Markeesha sold 30 boxes of crackers on Friday. On Saturday, she sold twice as many as on Friday. So she sold 2 * 30 = 60 boxes. On Sunday, she sold 15 fewer than Saturday. So she sold 60 - 15 = 45 boxes. The answer is 45. \\ \hline
        \end{tabular}
        \caption{Examples of the question with student disagreement = 2 in GSM8k dataset}
    \end{subtable}
    \bigskip
    \begin{subtable}{1\textwidth}
    \centering
        \begin{tabular}{m{6em}|m{33em}}
            \hline
            \textbf{Question} & There are 30 pieces of popcorn in a serving. Jared can eat 90 pieces of popcorn and his three other friends can each eat 60 pieces of popcorn. How many servings of popcorn should Jared order for all of them? \\ \hline
            \textbf{Generation 1} & Let's think step by step. There are 30 pieces of popcorn in the serving. Jared can eat 90 pieces of popcorn. His three other friends can each eat 60 pieces of popcorn. So Jared needs to order 3 * 90 = 270 pieces of popcorn. The answer is 270. \\ \hline
            \textbf{Generation 2} & Let's think step by step. There are initially 30 pieces of popcorn in a serving. Jared can eat 90 pieces of popcorn and his three other friends can each eat 60 pieces of popcorn. So 90 + 60 = 150. Jared should order 150 pieces. Jared can eat 150 - 30 = 120 pieces of popcorn. His three friends can eat 60 pieces of popcorn. So Jared's three friends can each eat 90 - 60 = 30 pieces of popcorn. The answer is 30. \\ \hline
            \textbf{Generation 3} & Let's think step by step. Jared started with 30 pieces of popcorn, but he can eat 90 pieces of popcorn. 60 pieces each is 3 x 30 = 90 pieces. Jared has 90 - 60 pieces. Jared should order 30 servings of popcorn for all of the friends. The answer is 30. \\ \hline
            \textbf{Generation 4} & Let's think step by step. There are 30 pieces of popcorn. Jared can eat 90 pieces. The other three can eat 60 pieces. So Jared needs 90 + 60 = 150 pieces of popcorn. The answer is 150. \\ \hline
            \textbf{Generation 5} &Let's think step by step. Jared can eat 90 pieces. So he can eat 30 pieces a serving. His three friends can each eat 60 pieces. 60 pieces * 3 = 180 pieces. 30 pieces x 180 = 5400 pieces. Jared can order 5400 pieces for all of them. The answer is 5400. \\ \hline
        \end{tabular}
        \caption{Examples of the question with student disagreement = 4 in GSM8k dataset}
    \end{subtable}
    \caption{\label{table:disagreement_examples} Examples of questions as well as the student models' generations with different disagreement (student internal signal $s_i$).}

\end{table*}

\begin{table*}[ht]
    \begin{subtable}{1\textwidth}
    \centering
        \begin{tabular}{m{10em}|m{28em}}
            \hline
            \textbf{Question} & There are 290 liters of oil in 24 cans. If 10 of the cans are holding 8 liters each, how much oil is each of the remaining cans holding? \\ \hline
            \textbf{Teacher Generation} & Let's think step by step. There are 290 liters of oil in 24 cans. If 10 of the cans are holding 8 liters each, then the total amount of oil in those 10 cans is 10 x 8 = 80 liters. So the remaining oil in the other cans is 290 - 80 = 210 liters. Since there are 24 - 10 = 14 remaining cans, each of those cans is holding 210 / 14 = 15 liters. Therefore, each of the remaining cans is holding 15 liters of oil. \\ \hline
            \textbf{TA Confidence} & Confidence Choice: (a) very confident \\ \hline
        \end{tabular}
        \caption{Examples of the question, teacher annotation and TA confidence = ``very confident'' in GSM8k dataset}
    \end{subtable}
    \bigskip
    \begin{subtable}{1\textwidth}
    \centering
        \begin{tabular}{m{10em}|m{28em}}
            \hline
            \textbf{Question} & It is recommended that men should consume no more than 150 calories of added sugar per day. Mark took a soft drink in the afternoon that contained 2500 calories, 5\% of which was from added sugar. Then he bought some bars of candy which had 25 calories of added sugar each. If he eventually exceeded the recommended intake of added sugar by 100\%, how many bars of candy did he take? \\ \hline
            \textbf{Teacher Generation} & Let's think step by step. First, we need to find out how many calories of added sugar Mark consumed from the soft drink. Since 5\% of the soft drink's calories were from added sugar, we can calculate 2500 * 0.05 = 125 calories of added sugar from the soft drink.
            \\ \hline
            \textbf{TA Confidence} & Confidence Choice: (d) wrong answer \\ \hline
        \end{tabular}
        \caption{Examples of the question, teacher annotation and TA confidence = ``wrong answer'' in GSM8k dataset}
    \end{subtable}
    \caption{\label{table:confidence_examples} Examples of questions as well as the teacher models' generations with the corresponding TA confidence (TA-teacher signal $s_e$).}
\end{table*}